\documentclass[11pt]{article}

\pdfoutput=1

\usepackage[final]{acl}

\usepackage{times}
\usepackage{latexsym}

\usepackage[T1]{fontenc}
\usepackage{tabularray}
\usepackage[utf8]{inputenc}

\usepackage{microtype}

\usepackage{inconsolata}

\usepackage{graphicx}

\usepackage{CJKutf8} 

\usepackage{caption}

\usepackage{tabularx}

\title{Distinct social-linguistic processing between humans and large audio-language models: Evidence from model-brain alignment}

\author{
\textbf{Hanlin Wu\textsuperscript{1}}\quad
\textbf{Xufeng Duan\textsuperscript{1}}\quad
\textbf{Zhenguang G. Cai\textsuperscript{1,2}}\\
\textsuperscript{1}Department of Linguistics and Modern Languages, The Chinese University of Hong Kong\\
\textsuperscript{2}Brain and Mind Institute, The Chinese University of Hong Kong\\
\texttt{hanlin.wu@link.cuhk.edu.hk}
}

\begin{document}
\begin{CJK*}{UTF8}{gbsn}

\maketitle
\begin{abstract}
Voice-based AI development faces unique challenges in processing both linguistic and paralinguistic information. This study compares how large audio-language models (LALMs) and humans integrate speaker characteristics during speech comprehension, asking whether LALMs process speaker-contextualized language in ways that parallel human cognitive mechanisms. We compared two LALMs' (Qwen2-Audio and Ultravox 0.5) processing patterns with human EEG responses. Using surprisal and entropy metrics from the models, we analyzed their sensitivity to speaker-content incongruency across social stereotype violations (e.g., a man claiming to regularly get manicures) and biological knowledge violations (e.g., a man claiming to be pregnant). Results revealed that Qwen2-Audio exhibited increased surprisal for speaker-incongruent content and its surprisal values significantly predicted human N400 responses, while Ultravox 0.5 showed limited sensitivity to speaker characteristics. Importantly, neither model replicated the human-like processing distinction between social violations (eliciting N400 effects) and biological violations (eliciting P600 effects). These findings reveal both the potential and limitations of current LALMs in processing speaker-contextualized language, and suggest differences in social-linguistic processing mechanisms between humans and LALMs.
\end{abstract}

\section{Introduction}

Humans are remarkably adept at extracting speaker characteristics from vocal cues. Within milliseconds of hearing a voice, listeners can perceive a speaker's gender, age, health condition, personality traits, and other socio-demographic attributes \citep{lavan2024time}. The perceived speaker attributes then form a critical context for language comprehension, shaping how linguistic input is processed and interpreted \citep{wu2024speaker}. For example, when we hear someone say “The first time I got \textit{pregnant} I had a hard time,” it is straightforward when coming from a female speaker but would be puzzling if a man were to say it.

Electroencephalography (EEG) studies show that when people hear sentences containing speaker incongruencies—such as “The first time I got \textit{pregnant}…” spoken by a man (violating biological knowledge) or “I like to get \textit{manicures}…” spoken by a man (violating gender stereotypes)—their brain responses diverge from speaker-congruent conditions, showing an N400 effect \citep{martin2016holiday,van2008neural,van2012empathy} or a P600 effect \citep{lattner2003talker,foucart2015does}. These neural responses show that speaker characteristics actively shape the real-time processing of spoken language.

The human capacity for speaker-contextualized language processing has recently been explained through a rational inference framework \citep{wu2024man}. This framework proposes that humans engage in rational inference during real-time language comprehension—a process where listeners actively reason about the most likely interpretation given both linguistic input and speaker characteristics. Using social-stereotype violation (e.g., men getting manicures) and biological-knowledge violation (e.g., men getting pregnant) as test cases, they showed that when encountering speaker-content mismatches that violate social stereotypes, listeners can still arrive at a “literal” interpretation through effortful integration with their social knowledge, reflected in N400 effects. However, when faced with biological impossibilities, listeners rationally infer potential errors in the input and engage in error correction processes, manifested as P600 effects.

Recent advances in large language models (LLMs) have demonstrated increasing capabilities in contextual understanding \citep{zhu2024can} and multimodal processing \citep{wang2024multimodal,zhang2024mm}. While initially focused on text, these models have expanded into multimodal tasks, showing remarkable abilities in integrating inputs from diverse modalities like vision and speech. This evolution has led to the development of large audio-language models (LALMs) that can process audio inputs, including speaker characteristics, acoustic features, along with other contextual information.

The integration of LLMs into audio processing has progressed through several stages \citep{peng2024survey}. Early attempts focused on incorporating Transformer architectures into traditional speech models, as exemplified by HuBERT's self-supervised learning on unlabeled speech data \citep{hsu2021hubert}. More recent approaches have shifted toward direct audio processing with LLMs by mapping audio features to tokens, not only for higher computational efficiency but also enabling richer paralinguistic processing through end-to-end multimodal integration (e.g., \citealp{chu2024qwen2}).

This paradigm shift has produced models that are capable of increasingly complex tasks: AudioPaLM can preserve speaker voice characteristics during speech processing and generation \citep{rubenstein2023audiopalm}, SALMONN can perform audio-based storytelling and speech-audio co-reasoning \citep{tang2023salmonn}, and Qwen2-Audio can explicitly identify speaker demographics and emotions \citep{chu2024qwen2}. These emerging abilities raise questions about whether LALMs process speaker-contextualized language in ways that parallel human cognitive mechanisms.

As these models are increasingly deployed in interactive settings where they must interpret and respond to diverse speakers, understanding their social-linguistic processing has both theoretical and practical implications. On the one hand, comparing LALMs with human processing can provide insights into models' emergent cognitive mechanisms, an approach that has been widely used with deep neural networks \citep{alkhamissi2024llm,alkhamissi2025language,schrimpf2018brain}; on the other hand, identifying divergences between human and model processing helps pinpoint potential limitations in current architectures or training method, suggesting directions for developing more natural human-AI interactions.

To this end, we utilize computational metrics that have been shown to capture humans' real-time language processing. Specifically, surprisal \citep{hale2001probabilistic,levy2008expectation}, which reflects the unpredictability of a word given its context, has been linked to increased processing effort and has been shown to predict reading times \citep{smith2013effect} and N400 amplitudes \citep{krieger2024limits,salicchi2025not}. Entropy, which captures the uncertainty within the probability distribution of upcoming stimuli, was suggested to be associated with P600 amplitudes \citep{salicchi2025not}.

In this research, we investigate whether LALMs align with human cognitive mechanisms in social-linguistic processing. We use the EEG data from \citet{wu2024man} as a benchmark of human processing and examine: a) whether LALMs align with humans in perceiving speaker characteristics and use them to guide real-time language processing; b) whether LALMs align with humans in the specific mechanism in processing speaker-content relationships.

\section{Method}

\subsection{Human EEG data}

The human data were EEG responses to speech stimuli from native Mandarin Chinese speakers. The study employed a 2×2 factorial design crossing Congruency (speaker-congruent vs. speaker-incongruent) with Type (social vs. biological). Congruency was manipulated by matching or mismatching speaker characteristics with the sentence content, while Type distinguished between violations of social stereotypes and biological knowledge. The experimental materials consisted of 80 self-referential sentences (each with a speaker-congruent and a speaker-incongruent audio version) in Mandarin Chinese, with speaker characteristics varying along gender and age dimensions (Table~\ref{item-examples}). All sentence audios were generated using text-to-speech technique with consistent acoustic properties.

\begin{table*}
  \centering
  \small
  \begin{tabularx}{\textwidth}{lXX}
    \hline
    \textbf{Category} & \textbf{Example} & \textbf{English translation}\\
    \hline
    SM & 在工作单位我一般都是穿\underline{西服}打领带。 & At the workplace I usually wear a \underline{suit} and a tie.\\
    SF & 这个周末我要先去做\underline{美甲}然后理发。 & This weekend I'm going to get a \underline{manicure} and then a haircut.\\
    SA & 我最近\underline{上班}压力太大需要休息。 & I've been \underline{working} too hard lately and I need a break.\\
    SC & 他把我的\underline{玩具}抢走了我要去找妈妈告状。 & He took my \underline{toys} away from me and I'm going to tell mummy about it.\\
    BM & 我需要定期去医院检查\underline{前列腺}的健康状况。 & I need to go to the hospital to check my \underline{prostate} on a regular basis.\\
    BF & 我第一次\underline{怀孕}的时候过得很艰难。 & The first time I got \underline{pregnant} I had a hard time.\\
    BA & 我发现我脸上的\underline{老年斑}越来越多了我正在寻找新的治疗方法。 & I noticed that I'm getting more and more \underline{age spots} on my face and I am looking for new treatments.\\
    BC & 我在等我的\underline{乳牙}掉下来然后我要把它扔到房顶上。 & I'm waiting for my \underline{milk tooth} to fall out and then I'm going to throw it on the roof.\\
    \hline
  \end{tabularx}
  \caption{\label{item-examples}
    Examples of Stimuli used in \citet{wu2024man} with English translations. SM: socially congruent with male speakers; SF: socially congruent with female speakers; SA: socially congruent with adult speakers; SC: socially congruent with child speakers; BM: biologically congruent with male speakers; BF: biologically congruent with female speakers; BA: biologically congruent with adult speakers; BC: biologically congruent with child speakers. Critical words are underscored.
  }
\end{table*}

The EEG data were collected from 60 participants while they listened to these sentences. A region of interest of 59 central-posterior sites was selected, and trial-level amplitudes were averaged across these sites before being further averaged over 300-600 ms (N400) and 600-1000 ms (P600) post-critical word onset. Their results revealed that social incongruency elicited a long-lasting N400 effect (across the 300-600-ms and the 600-1000-ms time windows), while biological incongruency elicited a P600 effect (600-1000 ms).

\subsection{LALM metrics}
We collected the computational metrics from two LALMs: Qwen2-Audio 7B Instruct \citep{chu2024qwen2} and Ultravox 0.5 8B (www.ultravox.ai). We obtained the surprisal and entropy of the critical word through a sentence continuation task where we inputted the audio sentences that were cut short at the critical word following a text-based instruction to guide the model to continue the audio sentence by outputting text (see Appendix for prompts). 

Surprisal was computed as the negative log probability of the target word given its context:

\begin{equation}
S(w_t) = -\log_2 P(w_t|C)
\end{equation}

Where $w_t$ represents the target word (i.e., the critical word that distinguishes speaker-congruent and -incongruent conditions); $C$ represents the context before the target word, including the text-based instruction and the audio sentence; $P(w_t|C)$ was the word probability. For words containing multiple tokens, we calculated the joint probability at the token level.

Entropy was calculated over the probability distribution of the model's predictions at the target word position:

\begin{equation}
H(w_t) = - \sum P(w_x|C) \log_2 P(w_x|C)
\end{equation}

Where $w_x$ represents possible continuations. For words containing multiple tokens, we calculated the sum of the entropy for each token in the word. To test the generalizability across languages, we additionally created an English version of each sentence by translation and adaptation. The English audio was generated using the same standard as the Chinese audio. Metrics were collected for both the original Chinese stimuli and their English translations to test cross-linguistic generalization. We also collected these metrics from the text-based stimuli (the text transcription of those audio sentences) to serve as the baseline.

\section{Results}

We examined the model-brain alignment from two perspectives. First, we examined whether the LALM response patterns resembled humans by replicating the analyses in the human study on LALM data. Second, we examined whether LALM responses could predict human brain responses by including LALM metrics as additional predictors for the human brain responses. For all analyses, we used linear mixed-effects (LME) modeling with maximal random-effect structure determined by forward model comparison ($\alpha$ = 0.2, \citealp{matuschek2017balancing}). For surprisal and entropy analyses, we used item-level data and included the random effect of Item; for model-EEG alignment analyses, we used trial-level data and included the random effects of both Participant and Item.

\begin{figure}[t]
  \includegraphics[width=\columnwidth]{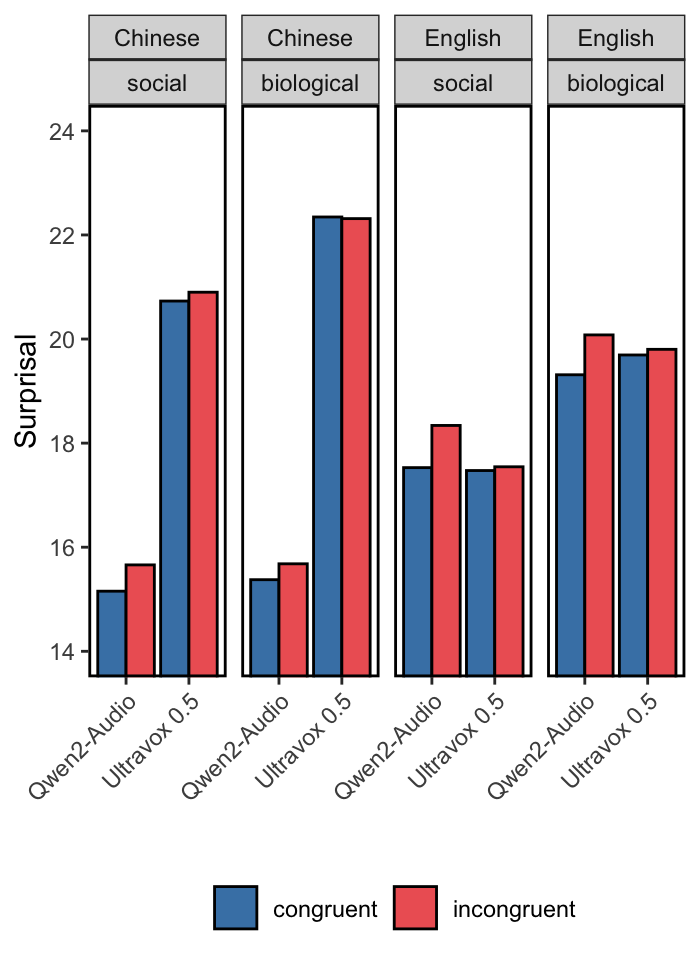}
  \caption{Surprisal values from Qwen2-Audio and Ultravox 0.5 models for speaker-congruent (blue) and speaker-incongruent (red) utterances, shown separately for social and biological conditions in Chinese and English.}
  \label{fig:experiments}
\end{figure}

\subsection{Surprisal (Qwen2-Audio)}

To test whether surprisal metric replicated the human brain pattern, we conducted LME analyses with Congruency (congruent = -0.5, incongruent = 0.5) and Type (social = -0.5, biological = 0.5) as interacting fixed effects, along with text-based surprisal as control, and showed a significant main effect of Congruency ($\beta$ = 0.41, $SE$ = 0.19, $t$ = 2.12, $p$ = .037) and text-based surprisal ($\beta$ = 3.97, $SE$ = 0.30, $t$ = 13.17, $p$ < .001), suggesting that the LALM model was sensitive to speaker-content incongruency regardless of violation type. The critical interaction between Congruency and Type was absent ($\beta$ = -0.20, $SE$ = 0.38, $t$ = -0.52, $p$ = .602), suggesting that unlike humans, the model processed social and biological violations similarly. The same pattern was replicated in English materials, as there was a significant main effect of Congruency ($\beta$ = 0.73, $SE$ = 0.20, $t$ = 3.55, $p$ < .001) and text-based surprisal ($\beta$ = 3.89, $SE$ = 0.32, $t$ = 12.14, $p$ < .001), while the interaction between Congruency and Type was absent ($\beta$ = -0.17, $SE$ = 0.41, $t$ = -0.42, $p$ = .678).

\subsection{Surprisal (Ultravox 0.5)}

Unlike Qwen2-Audio, the results for Ultravox 0.5 only showed a significant main effect of text-based surprisal ($\beta$ = 4.09, $SE$ = 0.36, $t$ = 11.32, $p$ < .001) for Chinese materials, while the main effect of Congruency ($\beta$ = 0.07, $SE$ = 0.08, $t$ = 0.85, $p$ = .399) or the interaction between Congruency and Type was absent ($\beta$ = -0.20, $SE$ = 0.16, $t$ = -1.23, $p$ = .222), suggesting that this model might not be sensitive to speaker-content relationships. The same pattern was shown in English materials, as there was a significant main effect of text-based surprisal ($\beta$ = 2.41, $SE$ = 0.42, $t$ = 5.71, $p$ < .001), and no main effect of Congruency ($\beta$ = 0.05, $SE$ = 0.10, $t$ = 0.44, $p$ = .663) or interaction between Congruency and Type ($\beta$ = -0.05, $SE$ = 0.21, $t$ = -0.26, $p$ = .799).

\subsection{Entropy (Qwen2-Audio)}

To test whether entropy metric replicated the human brain pattern, we conducted LME analyses with Congruency and Type as interacting fixed effects, along with text-based entropy as control, and showed that there was only a significant main effect of text-based entropy ($\beta$ = 10.27, $SE$ = 0.36, $t$ = 28.49, $p$ < .001) for Chinese materials. Neither the main effect of Congruency ($\beta$ = -0.09, $SE$ = 0.15, $t$ = -0.61, $p$ = .546) nor the interaction between Congruency and Type ($\beta$ = -0.17, $SE$ = 0.30, $t$ = -0.55, $p$ = .582) reached significance, suggesting that the model's uncertainty in prediction was primarily driven by the linguistic properties of the input rather than speaker-content relationships. The same pattern was shown in English materials, as the main effect of text-based entropy emerged ($\beta$ = 8.71, $SE$ = 0.42, $t$ = 20.89, $p$ < .001), while the main effect of Congruency ($\beta$ = -0.04, $SE$ = 0.17, $t$ = -0.21, $p$ = .836) and the interaction between Congruency and Type remained absent ($\beta$ = -0.37, $SE$ = 0.34, $t$ = -1.07, $p$ = .289).

\subsection{Entropy (Ultravox 0.5)}

The pattern observed in Qwen2-Audio was replicated with Ultravox 0.5, as there was only a significant main effect of text-based entropy ($\beta$ = 9.50, $SE$ = 0.23, $t$ = 41.77, $p$ < .001) for Chinese materials. Neither the main effect of Congruency ($\beta$ = -0.01, $SE$ = 0.02, $t$ = -0.48, $p$ = .630) nor the interaction between Congruency and Type ($\beta$ = 0.04, $SE$ = 0.05, $t$ = 0.90, $p$ = .369) reached significance. This pattern was further replicated with English materials, as there was only a significant main effect of text-based entropy ($\beta$ = 3.63, $SE$ = 0.70, $t$ = 5.22, $p$ < .001). Neither the main effect of Congruency ($\beta$ = -0.03, $SE$ = 0.04, $t$ = -0.64, $p$ = .526) nor the interaction between Congruency and Type ($\beta$ = -0.05, $SE$ = 0.08, $t$ = -0.63, $p$ = .531) reached significance.

\begin{figure}[t]
  \includegraphics[width=\columnwidth]{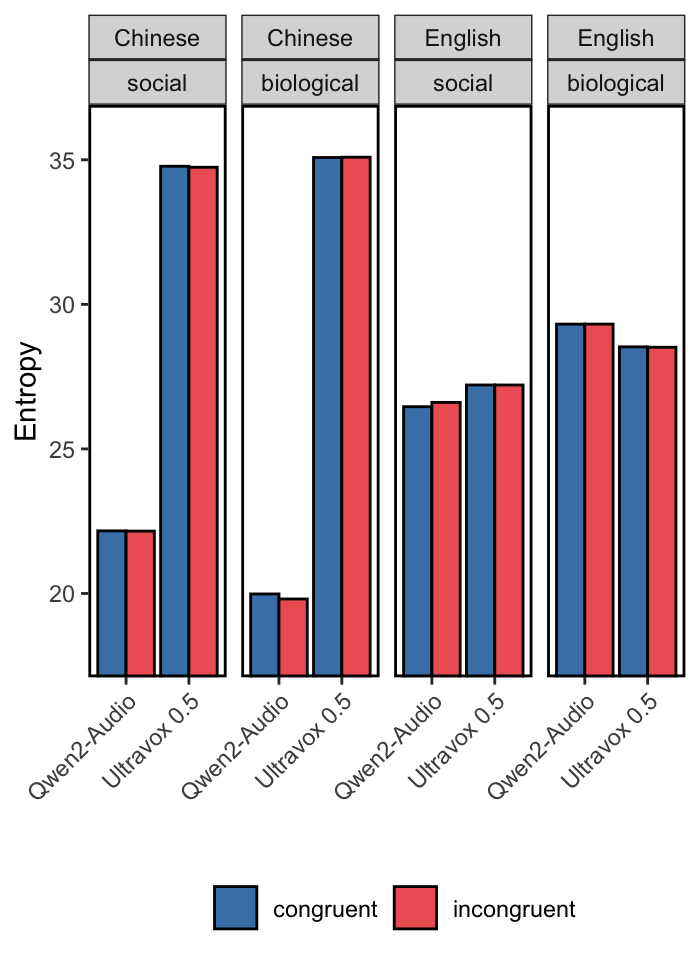}
  \caption{Entropy values from Qwen2-Audio and Ultravox 0.5 models for speaker-congruent (blue) and speaker-incongruent (red) utterances, shown separately for social and biological conditions in Chinese and English.}
  \label{fig:experiments}
\end{figure}

\subsection{Surprisal-EEG alignment (Qwen2-Audio)}

To test whether Surprisal could predict EEG response and whether the prediction varied across conditions, we added Surprisal (a scaled continuous variable) as a fixed effect interacting with Congruency and Type to the original LME analyses of EEG amplitude in Wu and Cai (2024a). For 300-600 ms, the results revealed a significant main effect of Surprisal ($\beta$ = -0.50, $SE$ = 0.16, $t$ = -3.12, $p$ = .002), while it did not interact with Congruency ($\beta$ = 0.15, $SE$ = 0.26, $t$ = 0.57, $p$ = .574), Type ($\beta$ = 0.18, $SE$ = 0.32, $t$ = 0.56, $p$ = .579), or the interaction between Congruency and Type ($\beta$ = 0.17, $SE$ = 0.52, $t$ = 0.34, $p$ = .736). For 600-1000 ms, there was no main effect of Surprisal ($\beta$ = -0.22, $SE$ = 0.18, $t$ = -1.21, $p$ = .229), or interaction with Congruency ($\beta$ = 0.23, $SE$ = 0.30, $t$ = 0.76, $p$ = .447), Type ($\beta$ = 0.53, $SE$ = 0.36, $t$ = 1.50, $p$ = .138), or three-way interaction with Congruency and Type ($\beta$ = -0.09, $SE$ = 0.60, $t$ = -0.16, $p$ = .877). These results suggested that surprisal significantly predicted N400 responses in a condition-independent manner, while it did not contribute to P600 responses.

\subsection{Surprisal-EEG alignment (Ultravox 0.5)}

For 300-600 ms, the results revealed a marginally significant main effect of Surprisal ($\beta$ = -0.33, $SE$ = 0.18, $t$ = -1.79, $p$ = .078), while it did not interact with Congruency ($\beta$ = 0.33, $SE$ = 0.28, $t$ = 1.16, $p$ = .250), Type ($\beta$ = 0.23, $SE$ = 0.37, $t$ = 0.62, $p$ = .539), or the interaction between Congruency and Type ($\beta$ = -0.34, $SE$ = 0.56, $t$ = -0.60, $p$ = .548). For 600-1000 ms, there was no main effect of Surprisal ($\beta$ = 0.26, $SE$ = 0.20, $t$ = 1.26, $p$ = .212), or its interaction with Congruency ($\beta$ = 0.19, $SE$ = 0.32, $t$ = 0.61, $p$ = .544), Type ($\beta$ = 0.46, $SE$ = 0.38, $t$ = 1.22, $p$ = .227), or the three-way interaction with Congruency and Type ($\beta$ = -0.31, $SE$ = 0.63, $t$ = -0.49, $p$ = .626). These results suggested that unlike Qwen2-Audio, Ultravox 0.5's surprisal did not reliably predict either N400 or P600 responses, despite showing a trend predicting N400.

\subsection{Entropy-EEG alignment (Qwen2-Audio)}

To test whether Entropy can predict EEG response and whether the prediction varied across conditions, we added Entropy (a scaled continuous variable) as a fixed effect interacting with Congruency and Type. The results revealed no significant main effect of Entropy (300-600 ms: $\beta$ = -0.19, $SE$ = 0.17, $t$ = -1.14, $p$ = .259; 600-1000 ms: $\beta$ = 0.02, $SE$ = 0.18, $t$ = 0.12, $p$ = .907), or interaction with Congruency (300-600 ms: $\beta$ = 0.34, $SE$ = 0.26, $t$ = 1.31, $p$ = .193; 600-1000 ms: $\beta$ = -0.05, $SE$ = 0.30, $t$ = -0.18, $p$ = .861), Type (300-600 ms: $\beta$ = 0.44, $SE$ = 0.34, $t$ = 1.29, $p$ = .202; 600-1000 ms: $\beta$ = 0.36, $SE$ = 0.37, $t$ = 0.98, $p$ = .333), or the three-way interaction with Congruency and Type (300-600 ms: $\beta$ = -0.64, $SE$ = 0.52, $t$ = -1.24, $p$ = .220; 600-1000 ms: $\beta$ = -0.59, $SE$ = 0.60, $t$ = -0.98, $p$ = .329). These results suggested that the model's predictive uncertainty did not predict human neural responses for either N400 or P600.

\subsection{Entropy-EEG alignment (Ultravox 0.5)}

The results revealed no significant main effect of Entropy in the N400 time window (300-600 ms: $\beta$ = 0.04, $SE$ = 0.18, $t$ = 0.20, $p$ = .844), but a marginal main effect in the P600 time window (600-1000 ms: $\beta$ = 0.31, $SE$ = 0.18, $t$ = 1.76, $p$ = .083). There were no significant interactions with Congruency (300-600 ms: $\beta$ = 0.32, $SE$ = 0.25, $t$ = 1.28, $p$ = .204; 600-1000 ms: $\beta$ = 0.27, $SE$ = 0.30, $t$ = 0.89, $p$ = .375), Type (300-600 ms: $\beta$ = 0.11, $SE$ = 0.34, $t$ = 0.33, $p$ = .744; 600-1000 ms: $\beta$ = 0.17, $SE$ = 0.36, $t$ = 0.48, $p$ = .633), or the three-way interaction with Congruency and Type (300-600 ms: $\beta$ = -0.48, $SE$ = 0.50, $t$ = -0.96, $p$ = .340; 600-1000 ms: $\beta$ = -0.27, $SE$ = 0.60, $t$ = -0.45, $p$ = .652). These results suggested that, similar to Qwen2-Audio, the model's predictive uncertainty did not strongly predict human neural responses for either N400 or P600, though there was a trend for higher entropy to predict larger P600 amplitudes.

\begin{figure}[t]
  \includegraphics[width=\columnwidth]{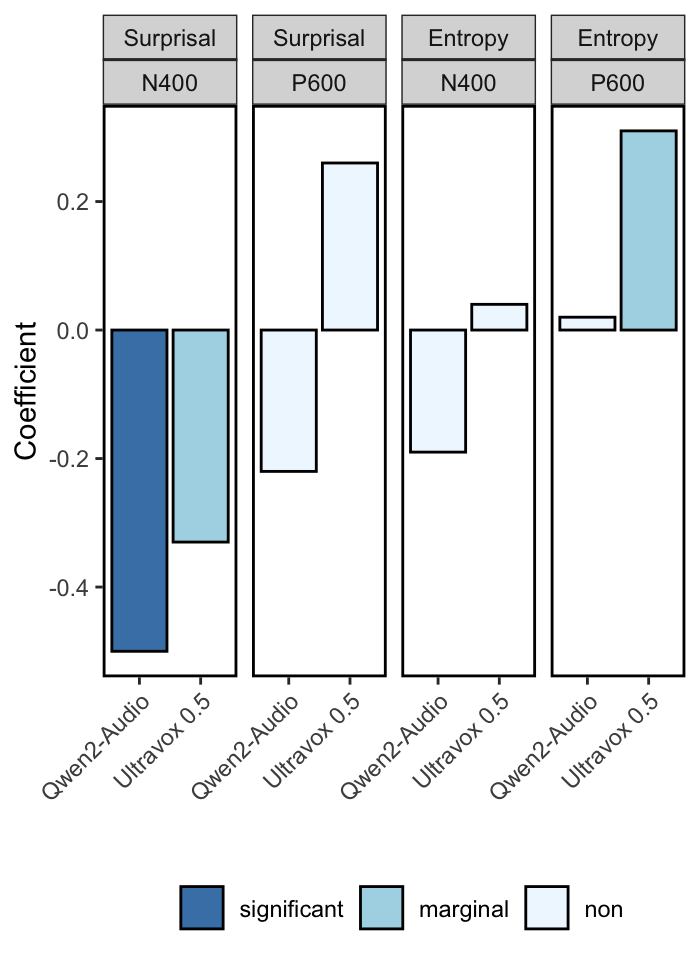}
  \caption{Main effect coefficients of Surprisal and Entropy on N400 and P600 amplitudes from LME analyses. Dark blue indicates a significant effect, light blue indicates marginal effects, and alice blue indicates non-significant effects for Qwen2-Audio and Ultravox 0.5 models.}
  \label{fig:experiments}
\end{figure}

\section{Discussion}

Our results revealed varying degrees of alignment between humans and LALMs in the social-linguistic processing of speech. Qwen2-Audio showed sensitivity to speaker-content incongruency through increased surprisal for incongruent utterances and significantly predicted human N400 responses. In contrast, Ultravox 0.5 showed no sensitivity to speaker-content relationships in its surprisal patterns and did not reliably predict human neural responses, despite showing a trend for N400.

Moreover, neither model showed human-like distinctions between social and biological violations, and both models' predictive uncertainty (entropy) was primarily driven by linguistic properties rather than speaker-content relationships and generally did not predict human neural responses, though Ultravox 0.5 showed a marginal trend for higher entropy predicting larger P600 amplitudes.

The distinct neural signatures for social versus biological violations in humans likely reflect different cognitive mechanisms. As \citet{wu2024man} suggested, social violations may be processed through semantic integration where linguistic content and speaker characteristics are integrated with prior knowledge about social roles and stereotypical behaviors, leading to N400 effects. In contrast, biological violations may trigger error detection and correction processes that attempt to resolve the physical impossibility, resulting in P600 effects. This distinction reflects rationality in human cognition.

Unlike humans who engage in active reanalysis when encountering biological impossibilities (reflected in the P600), current LALMs operate through single-pass forward prediction without mechanisms for backtracking or reanalysis. This may relate to the fact that current LLMs are typically trained to predict tokens one at a time, optimizing for local coherence rather than longer-range consistency. While some models are beginning to explore multi-token prediction windows \citep{gloeckle2024better,liu2024deepseek} that could theoretically capture longer-range dependencies and support reanalysis-like processes, most still lack similar mechanisms.

An open question is the precise mechanism by which LALMs utilize speaker information in their predictions. Unlike humans who readily identify speaker characteristics from voice and use this information to guide comprehension, it remains unclear whether LALMs explicitly represent speaker identity (e.g., assigning a gender category to a voice) or simply learn statistical associations between acoustic features and linguistic content. This distinction has implications for understanding both model processing and human cognition. For humans, the N400 and P600 effects depend on correctly identifying speaker characteristics and applying relevant world knowledge. If LALMs do not explicitly represent speaker identity but still show some degree of sensitivity to speaker-content relationships, this would suggest that explicit categorization may not be necessary for content prediction, though it might be essential for the rational inference processes that humans employ when resolving incongruencies. Future research could probe this question by examining model representations of speaker characteristics and their relationship to linguistic predictions.

Lastly, our findings also raise ethical considerations regarding LALMs' gender (and age) bias, which has been widely shown in LLMs \citep{kotek2023gender,zhao2024gender}. The observation that Qwen2-Audio showed increased surprisal for gender-nonconforming utterances indicates that it might have internalized societal gender stereotypes during training. While such sensitivity may facilitate natural interactions with humans, it also risks perpetuating harmful stereotypes if deployed in applications that influence decision-making or content generation.

In conclusion, we show that LALMs can potentially detect speaker-content violations and predict human N400 responses, but this capability varies between models. While Qwen2-Audio showed some level of alignment with human processing, neither Qwen2-Audio nor Ultravox 0.5 captured the human-like rational inference (as reflected by the distinction between social and biological violations), suggesting potential limitations in current LALM architectures or LLM architectures in general regarding real-time error analysis mechanisms.

\section{Limitations}

Several limitations of the current study should be acknowledged. First, our analyses focused on only two LALMs with a relatively small set of stimuli, which may not be representative of all current audio-language models or the full range of potential speaker-content relationships. A larger-scale investigation would better characterize the variation in speaker-content processing capabilities across different model architectures and training paradigms. Additionally, While surprisal and entropy are established metrics that have been linked to N400 and P600 responses respectively, they may be insufficient to capture the full range of processing distinctions that humans exhibit. Future research could explore alternative metrics such as analyzing activation patterns in different model layers, or utilizing representation similarity analysis between model embeddings and neural data. Finally, we only examined models' "static" responses to speaker characteristics, whereas humans show dynamic adaptation to individual speakers over increasing contexts \citep{grant2020male}. Human listeners rapidly adjust their predictions based on a speaker's established patterns—for example, becoming less surprised by stereotype-incongruent statements from a speaker who consistently violates stereotypes. This adaptive processing, which involves updating speaker models in real-time and adjusting predictions accordingly \citep{wu2025probabilistic}, represents an aspect of human language processing that our current single-utterance design cannot capture. Future work should examine how LALMs' predictions evolve across multiple utterances from the same speaker to better assess their capability for speaker-specific adaptation.

\end{CJK*}
\bibliography{acl_latex_hanlin}

\appendix

\section{Appendix: prompts for sentence continuation task}
\begin{CJK*}{UTF8}{gbsn}
\label{sec:appendix}

\subsection{Qwen2-Audio}
\subsection*{Chinese materials (audio)}
System: 你是一个实验中的参与者，你需要仔细听下面的录音。

User: 请补全录音中的句子，例如`我喜欢吃`，你可以回答`苹果`。直接回答补充的内容，不要说其他内容。录音：

User: (audio)

\subsection*{English materials (audio)}
System: You are a participant in an experiment, you need to listen carefully to the following recording.

User: Please complete the sentence from the recording, for example if you hear 'I like to eat', you can answer 'apples'. Just answer with the completing content, don't say anything else. Recording:

User: (audio)

\subsection*{Chinese materials (text)}
System: 你是一个实验中的参与者，你需要认真完成下面的任务。

User: 请补全以下句子。例如，`我喜欢吃`，你可以回答`苹果`，直接回答补充的内容，不要说其他内容。句子：(text)

\subsection*{English materials (text)}
System: You are a participant in an experiment, you need to complete the following task carefully.
User: Please complete the following sentence. For example, 'I like to eat', you can answer 'apples'. Just answer with the completing content, don't say anything else. Sentence: (text)

\subsection{Ultravox 0.5}
\subsection*{Chinese materials (audio)}
System: 请补全录音中的句子，例如 '我喜欢吃'，你可以回答 '苹果'。直接回答补充的内容，不要说其他内容。(audio)

\subsection*{English materials (audio)}
System: Please complete the sentence from the recording. For example, if you hear 'I like to eat', you can answer 'apples'. Just answer with the completing content, don't say anything else. (audio)

\subsection*{Chinese materials (text)}
System: 请补全以下句子。例如，'我喜欢吃'，你可以回答 '苹果'。直接回答补充的内容，不要说其他内容。句子：(text)

\subsection*{English materials (text)}
System: Please complete the following sentence. For example, if you hear 'I like to eat', you can answer 'apples'. Just answer with the completing content, don't say anything else. Sentence: (text)

\end{CJK*}
\end{document}